\documentclass[conference]{IEEEtran}

\usepackage{cite}

\ifCLASSINFOpdf
  \usepackage[pdftex]{graphicx}
  \graphicspath{{/pdf/}{/jpeg/}}
  \DeclareGraphicsExtensions{.pdf,.jpeg,.png}
\else
\fi
\usepackage{amsmath}
\usepackage{algorithm}
\usepackage[noend]{algpseudocode}	

\usepackage{array}

\usepackage{amssymb}	

\newcommand{\refequation}[1]{Equation~\ref{#1}}
\newcommand{\reffigure}[1]{Figure~\ref{#1}}
\newcommand{\refsection}[1]{Section~\ref{#1}}
\newcommand{\reftable}[1]{Table~\ref{#1}}

\newcommand*\mean[1]{%
  \hbox{%
    \vbox{%
      \hrule height 0.5pt 
      \kern0.4ex
      \hbox{%
        \kern-0.1em
        \ensuremath{#1}%
        \kern-0.1em
      }%
    }%
  }%
}

\usepackage{color}
\usepackage{makecell}

\begin{document}
\title{Enhancements for Real-Time Monte-Carlo Tree Search in General Video Game Playing}

\author{\IEEEauthorblockN{Dennis J. N. J. Soemers, Chiara F. Sironi, Torsten Schuster, and Mark H. M. Winands}
\IEEEauthorblockA{Department of Data Science and Knowledge Engineering, Maastricht University\\
d.soemers@gmail.com, t.schuster@student.maastrichtuniversity.nl, \{c.sironi,m.winands\}@maastrichtuniversity.nl}}

\maketitle

\begin{abstract}
General Video Game Playing (GVGP) is a field of Artificial Intelligence where agents play a variety of real-time video games that are unknown in advance. This limits the use of domain-specific heuristics. Monte-Carlo Tree Search (MCTS) is a search technique for game playing that does not rely on domain-specific knowledge. This paper discusses eight enhancements for MCTS in GVGP; Progressive History, N-Gram Selection Technique, Tree Reuse, Breadth-First Tree Initialization, Loss Avoidance, Novelty-Based Pruning, Knowledge-Based Evaluations, and Deterministic Game Detection. Some of these are known from existing literature, and are either extended or introduced in the context of GVGP, and some are novel enhancements for MCTS. Most enhancements are shown to provide statistically significant increases in win percentages when applied individually. When combined, they increase the average win percentage over sixty different games from 31.0\% to 48.4\% in comparison to a vanilla MCTS implementation, approaching a level that is competitive with the best agents of the GVG-AI competition in 2015.
\end{abstract}

\section{Introduction} \label{sec:Introduction}
General Video Game Playing (GVGP) \cite{Levine2013GVG} is a field of Artificial Intelligence in games where the goal is to develop agents that are able to play a variety of real-time video games that are unknown in advance. It is closely related to General Game Playing (GGP) \cite{Genesereth2005GGP}, which focuses on abstract games instead of video games. The wide variety of games in GGP and GVGP makes it difficult to use domain-specific knowledge, and promotes the use of generally applicable techniques.

There are two main frameworks for GVGP. The first framework is the Arcade Learning Environment (ALE) \cite{Bellemare2013ALE} for developing agents that can play games of the Atari 2600 console. The second framework is GVG-AI \cite{Perez2015GVG14}, which can run any real-time video game described in a Video Game Description Language \cite{Ebner2013TowardsVGDL,Schaul2013VGDL}. This paper focuses on the GVG-AI framework.

The GVG-AI framework is used in the GVG-AI Competition \cite{Perez2015GVG14,PerezGVG15}. Past competitions only ran a \textit{Planning Track}, where agents were ranked based on their performance in single-player games. In 2016, it is planned to extend this with a \textit{2/N-Player Track}, a \textit{Learning Track}, and a \textit{Procedural Content Generation Track}. This paper focuses on the Planning Track.

Monte-Carlo Tree Search (MCTS) \cite{Kocsis2006UCT,Coulom2007} is a popular technique in GGP \cite{Bjornsson2009CadiaPlayer} because it does not rely on domain-specific knowledge. MCTS has also performed well in GVGP in 2014 \cite{Perez2015GVG14}, which was the first year of the GVG-AI competition, but was less dominant in 2015 \cite{PerezGVG15}. This paper discusses and evaluates eight enhancements for MCTS to improve its performance in GVGP: \textit{Progressive History}, \textit{N-Gram Selection Technique}, \textit{Tree Reuse}, \textit{Breadth-First Tree Initialization}, \textit{Loss Avoidance}, \textit{Novelty-Based Pruning}, \textit{Knowledge-Based Evaluations} and \textit{Deterministic Game Detection}.

The remainder of the paper is structured as follows. \refsection{sec:GVGP} provides background information on the GVG-AI framework and the GVG-AI competition. MCTS is discussed in \refsection{sec:MCTS}. In \refsection{sec:Enhancements}, the enhancements for MCTS in GVGP are explained. \refsection{sec:Experiments} describes the experiments to assess the enhancements. Finally, the paper is concluded in \refsection{sec:Conclusion} and ideas for future research are discussed.

\section{GVG-AI Framework and Competition} \label{sec:GVGP}
In the GVG-AI competition \cite{Perez2015GVG14,PerezGVG15}, agents play a variety of games that are unknown in advance. Agents are given $1$ second of processing time at the start of every game, and $40$ milliseconds of processing time per \textit{tick}. A tick can be thought of as a turn in an abstract game. Every tick, the agent can choose an action to play, and at the end of the tick the chosen action is played and the game state progresses. Every game has a duration of at most $2000$ ticks, after which the game is a loss. Other than that, different games have different termination conditions, which define when the agent wins or loses. Every game in GVG-AI contains at least an \textit{avatar} object, which is the ``character'' controlled by the agent. Games can also contain many other types of objects. Games in GVG-AI are \textit{fully observable} and can be \textit{nondeterministic}.

Agents can perform searches and attempt to learn which actions are good using the \textit{Forward Model}, consisting of two important functions; \textit{advance} and \textit{copy}. Given a game state $s_t$, the \textit{advance($a$)} function can be used to generate a successor state $s_{t+1}$, which represents one of the possible states that can be reached by playing an action $a$. In deterministic games, there is only one such state $s_{t+1}$ for every action $a$, but in nondeterministic games there can be more than one. The \textit{copy($s_t$)} function creates a copy of $s_t$. This function is required when it is desirable to generate multiple possible successors of $s_t$, because every call to \textit{advance} modifies the original state, and there is no \textit{undo} function. Because the framework supports a wide variety of different games, it is not optimized as well as any framework dedicated to a specific game would be. This means that the \textit{advance} and \textit{copy} operations tend to be significantly slower than equivalent functions in individual game implementations.

\section{Monte-Carlo Tree Search} \label{sec:MCTS}
\begin{figure}[t]
\centering
\includegraphics[scale=0.5]{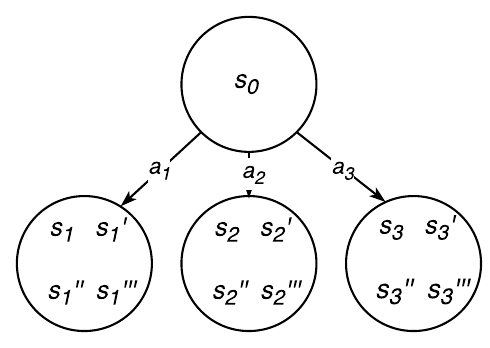}
\vspace{-12pt}
\caption{Example open-loop game tree. Nodes other than the root node can represent multiple possible states in nondeterministic games.}
\vspace{-15pt}
\label{fig:MCTSOpenLoopExample}
\end{figure}

Monte-Carlo Tree Search (MCTS) \cite{Kocsis2006UCT,Coulom2007} is a best-first search algorithm  that gradually builds up a search tree and uses Monte-Carlo simulations to approximate the value of game states. To handle nondeterministic games with probabilistic models that are not exposed to the agent, an \textit{``open-loop''} \cite{Perez2015OpenLoop} implementation of MCTS is used. In an open-loop approach, the root node represents the current game state ($s_0$), every edge represents an action, and every other node $n$ represents the set of game states that can be reached by playing the sequence of actions corresponding to the path from the root node to $n$, starting from $s_0$. See \reffigure{fig:MCTSOpenLoopExample} for an example.

\begin{figure}[t]
\centering
\includegraphics[scale=0.3]{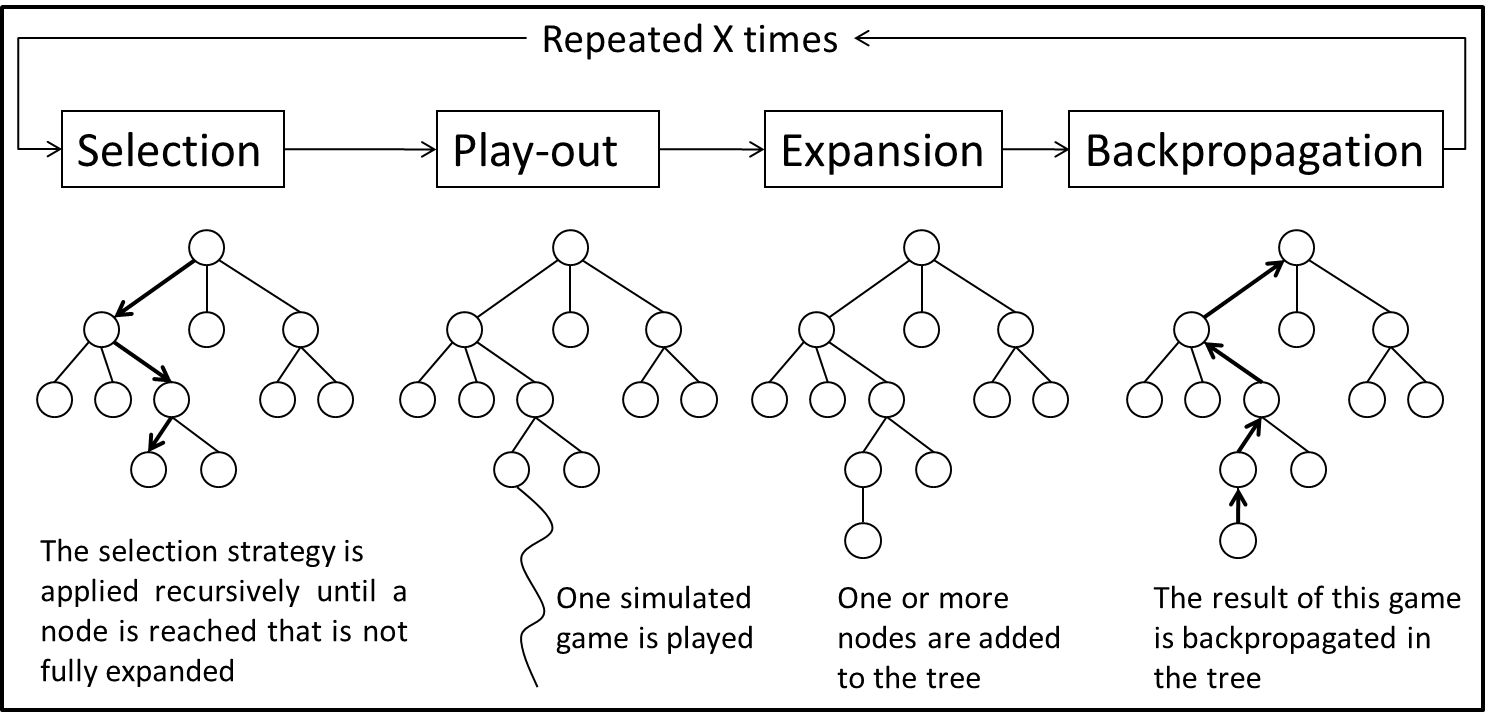}
\vspace{-5pt}
\caption{The four steps of an MCTS simulation. Adapted from \cite{Chaslot2008Progressive}.}
\vspace{-15pt}
\label{fig:MCTSSteps}
\end{figure}

MCTS is initialized with only the root node. Next, until some computational budget expires, the algorithm repeatedly executes simulations. Every simulation consists of the following four steps \cite{Chaslot2008Progressive}, depicted in \reffigure{fig:MCTSSteps}.

In the \textit{Selection} step, a \textit{selection policy} is applied recursively, starting from the root node, until a node is reached that is not yet fully expanded (meaning that it currently has fewer successors than available actions). The selection policy determines which part of the tree built up so far is evaluated in more detail. It should provide a balance between \textit{exploitation} of parts of the search tree that are estimated to have a high value so far, and \textit{exploration} of parts of the tree that have not yet been visited frequently. The most commonly implemented selection policy is \textit{UCB1} \cite{Kocsis2006UCT,Auer2002UCB}, which selects the successor $S_i$ of the current node $P$ that maximizes \refequation{Eq:UCT}. $S_i$ and $P$ are nodes, which can represent sets of states.
\begin{equation}
UCB1(S_i) = \mean{Q}(S_i) + C \times \sqrt{\frac{ln(n_P)}{n_i}}
\label{Eq:UCT}
\end{equation}
$\mean{Q}(S_i) \in [0, 1]$ denotes the normalized average score backpropagated through $S_i$ so far (as described below), $C$ is a parameter where higher values lead to more exploration, and $n_P$ and $n_i$ denote the visit counts of $P$ and $S_i$, respectively.

In the \textit{Play-out} step, the simulation is continued, starting from the last state encountered in the selection step, using a (semi-)random \textit{play-out policy}. The most straightforward implementation is to randomly draw actions to play from a uniform distribution until a terminal game state is reached. In GVGP, this is typically not feasible, and a maximum play-out depth is used to end play-outs early.

In the \textit{Expansion} step, the tree is expanded by adding one or more nodes. The most common implementation adds one node to the tree per simulation; the node corresponding to the first action played in the play-out step. In this paper, the tree is simply expanded by adding the whole play-out to the tree. The number of simulations per tick tends to be low enough in GVG-AI that there is no risk of running out of memory. Therefore, to keep all information gathered, all nodes are stored in memory.

In the \textit{Backpropagation} step, the outcome of the final state of the simulation is backpropagated through the tree. Let $s_T$ be the final state of the simulation. Next, an evaluation $X(s_T)$ of the state is added to a sum of scores stored in every node on the path from the root node to the final node of the simulation, and the visit counts of the same nodes are incremented. Because it is not feasible to let all simulations continue until terminal states are reached in GVG-AI, it is necessary to use some evaluation function for non-terminal states. A basic evaluation function that is also used by the sample MCTS controllers included in the GVG-AI framework is given by \refequation{Eq:GVGEval}.
\begin{equation}
X(s_T) = \begin{cases} 10^7 + score(s_T)
&\mbox{if } s_T $ is a winning state$ \\
-10^7 + score(s_T) & \mbox{if } s_T $ is a losing state$ \\
score(s_T) & \mbox{if } s_T $ is a non-terminal state$
\end{cases}
\label{Eq:GVGEval}
\end{equation}
$score(s_T)$ is the game score value of a state $s_T$ in GVG-AI. In some games a high game score value can indicate that the agent is playing well, but this is not guaranteed in all games.

Finally, the action leading to the node with the highest average score is played when the computational budget expires.

\section{MCTS Enhancements for GVGP} \label{sec:Enhancements}
There is a wide variety of existing enhancements for the MCTS algorithm, many of which are described in \cite{Browne2012SurveyMCTS}. This section discusses a number of enhancements that have been evaluated in GVGP; Progressive History, N-Gram Selection Technique, Tree Reuse, Breadth-First Tree Initialization, Loss Avoidance, Novelty-Based Pruning, Knowledge-Based Evaluations, and Deterministic Game Detection. Some are known from existing research, and some are new.

\subsection{Progressive History and N-Gram Selection Technique}
\textit{Progressive History} (PH) \cite{Nijssen2011Enhancements} and \textit{N-Gram Selection Technique} (NST) \cite{Tak2012NGram} are two existing enhancements for the selection and play-out steps of MCTS, respectively. The basic idea of PH and NST is to introduce a bias in the respective steps towards playing actions, or sequences of actions, that performed well in earlier simulations. Because the value of playing an action in GVG-AI typically depends greatly on the current position of the avatar, this position is also taken into account when storing data concerning the previous performance of actions. For a detailed description of these enhancements we refer to the original publications \cite{Nijssen2011Enhancements,Tak2012NGram}.

\subsection{Tree Reuse}
\begin{figure}[t]
\centering
\includegraphics[scale=0.45]{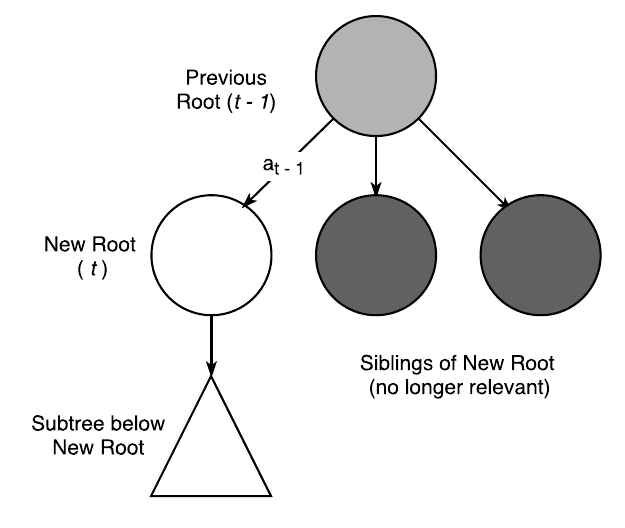}
\vspace{-15pt}
\caption{Tree Reuse in MCTS.}
\vspace{-15pt}
\label{fig:TreeReuse}
\end{figure}

Suppose that a search tree was built up by MCTS in a previous game tick $t - 1 \geq 0$, and an action $a_{t - 1}$ was played. The entire subtree rooted in the node corresponding to that action can still be considered to be relevant for the new search process in the current tick $t$. Therefore, instead of initializing MCTS with only a root node, it can be initialized with a part of the tree built in the previous tick, as depicted in \reffigure{fig:TreeReuse}. This was previously found to be useful in the real-time game of \textit{Ms Pac-Man} \cite{Pepels2014Pacman}. This idea has also previously been suggested in the context of GVGP \cite{Perez2015OpenLoop}, but, to the best of our knowledge, the effect of this enhancement on the performance of MCTS in GVGP has not yet been evaluated.

In nondeterministic games, it is possible that the new root (which was previously a direct successor of the previous root) represented more than one possible game state. In the current tick, it is known exactly which of those possible states has been reached. Therefore, some of the old results in this tree are no longer relevant. For this reason, all the scores and visit counts in the tree are decayed by multiplying them by a decay factor $\gamma \in [0, 1]$ before starting the next MCTS procedure. Tree Reuse (TR) with $\gamma=0$ completely resets the accumulated scores and visit counts of nodes (but still retains the nodes, and therefore the structure of the generated tree), and TR with $\gamma=1$ does not decay old results.

\subsection{Breadth-First Tree Initialization and Safety Prepruning}
In some of the games supported by the GVG-AI framework, the number of MCTS simulations that can be executed in a single tick can be very small; sometimes smaller than the number of available actions. In such a situation, MCTS behaves nearly randomly, and is susceptible to playing actions that lead to a direct loss, even when there are actions available that do not directly lose the game.

Theoretically this problem could be avoided by adjusting the limit of the play-out depth of MCTS to ensure that a sufficient number of simulations can be done. In practice, this can be problematic because it requires a low initial depth limit to ensure that it is not too high at the start of a game, and this can in turn be detrimental in games where it \textit{is} feasible and beneficial to run a larger number of longer play-outs.

We propose to handle this problem using Breadth-First Tree Initialization. The idea is straightforward; before starting MCTS, the direct successors of the root node are generated by a 1-ply Breadth-First Search. Every action available in the root state is executed up to a number $M$ times to deal with nondeterminism, and the resulting states are evaluated. The average of these $M$ evaluations is backpropagated for every successor with a weight equal to a single MCTS simulation. MCTS is only started after this process. When MCTS starts, every direct successor of the root node already has a prior evaluation that can be used to avoid playing randomly in cases with an extremely small number of simulations. The $M$ states generated for every successor are cached in the corresponding nodes, so that they can be re-used in the subsequent MCTS process. This reduces the computational overhead of the enhancement.

\textit{Safety prepruning}, originally used in an algorithm called Iterated Width \cite{Geffner2015GVGIW}, has been integrated in this process. The idea of safety prepruning is to count the number of immediate game losses among the $M$ generated states for each action, and only keep the actions leading to nodes with the minimum observed number of losses. All other actions are pruned.

\subsection{Loss Avoidance}
In GVGP, many games have a high number of losing game states that are relatively easy to avoid. An example of such a game is \textit{Frogs}, where the avatar is a frog that should cross a road and a river. The road contains trucks that cause a loss upon collision, but can easily be avoided because they move at a constant speed. The river contains logs that also move at a constant speed, which the frog should jump on in order to safely cross the river.

\begin{figure}[t]
\centering
\includegraphics[scale=0.4]{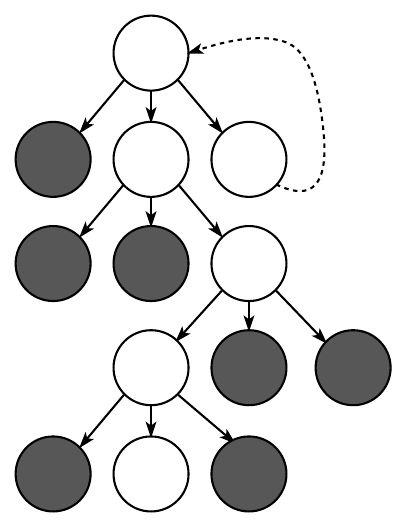}
\vspace{-5pt}
\caption{Example search tree. Dark nodes represent losing game states, and white nodes represent winning or neutral game states.}
\vspace{-15pt}
\label{fig:HighLossDensity}
\end{figure}

An example of a search tree with many losing states is depicted in \reffigure{fig:HighLossDensity}. In this example, the rightmost action in the root node is an action that brings the agent back to a similar state as in the root node. In the \textit{Frogs} game, this could be an action where the frog stays close to the initial position, and does not move towards the road. 

The (semi-)random play used in the play-out step of MCTS is likely to frequently run into losing game states in situations like this. This leads to a negative evaluation of nodes that do in fact lead to a winning position. This is only corrected when sufficient simulations have been run such that the selection step of MCTS correctly biases the majority of the simulations towards a winning node. With a low simulation count in GVG-AI, MCTS is likely to repeatedly play the rightmost action in \reffigure{fig:HighLossDensity}, which only delays the game until it is lost due to reaching the maximum game duration.

This problem is similar to the problem of \textit{traps} \cite{Ramanjuan2010Adversarial} or \textit{optimistic moves} \cite{Finnsson2011MCTSGameProperties} in (two-player) adversarial games. In those cases, MCTS has an overly optimistic evaluation of some states, whereas in the cases discussed here it has an overly pessimistic evaluation of some states. In \cite{Baier2015MctsMinimax}, it was proposed to integrate shallow minimax searches inside some of the steps of MCTS to improve its performance in game trees with \textit{traps} or \textit{optimistic moves}. Using minimax searches to \textit{prove} wins or losses is difficult in GVGP because games can be nondeterministic, but a similar idea can be used to get less pessimistic evaluations.

\begin{figure}[t]
\centering
\includegraphics[scale=0.5]{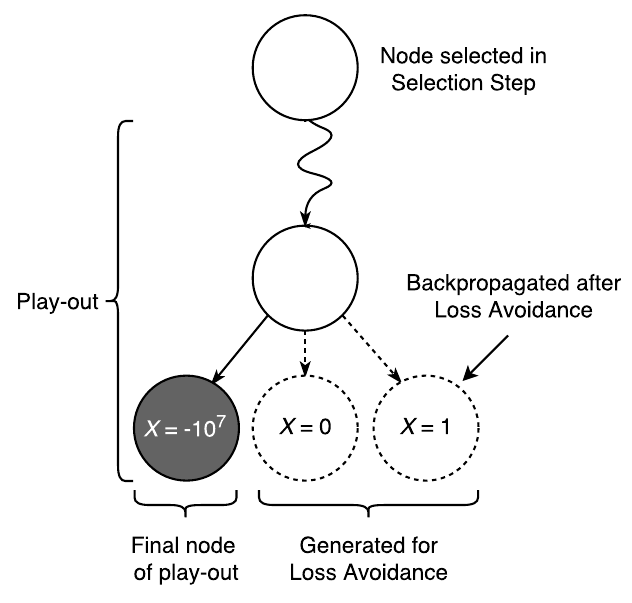}
\vspace{-10pt}
\caption{Example MCTS simulation with Loss Avoidance. The $X$ values in the last three nodes are evaluations of game states in those nodes. The dark node is a losing node.}
\vspace{-15pt}
\label{fig:LossAvoidance}
\end{figure}

In this paper, an idea named \textit{Loss Avoidance} (LA) is proposed for GVGP. The idea of LA is to try to ignore losses by immediately searching for a better alternative whenever a loss is encountered the first time a node is visited. An example is depicted in \reffigure{fig:LossAvoidance}. Whenever the play-out step of MCTS ends in a losing game state, that result is not backpropagated as would commonly be done in MCTS. Instead, one state is generated for every sibling of the last node, and only the evaluation of the node with the highest evaluation is backpropagated. All generated nodes are still added to the tree, and store their own evaluation in memory. 

LA causes MCTS to keep an optimistic initial view of the value of nodes. This tends to work well in the single-player games of GVG-AI, where it is often possible to reactively get out of dangerous situations. It is unlikely to work well in, for instance, adversarial games, where a high concentration of losses in a subtree typically indicates that an opposing player has more options to win and is likely in a stronger position.

In an open-loop implementation of MCTS, LA can have a significant amount of computational overhead in game trees with many losses. For instance, in the \textit{Frogs} game it roughly halves the average number of MCTS simulations per tick. This is because the node prior to the node with the losing game state does not store the corresponding game state in memory, which means that \textit{all} states generated in the selection and play-out steps need to be re-generated by playing the same action sequence from the root node. In nondeterministic games this process can also lead to finding a terminal state before the full action sequence has been executed again. To prevent spending too much time in the same simulation, the LA process is not started again, but the outcome of that state is backpropagated.

\subsection{Novelty-Based Pruning}
The concept of \textit{novelty tests} was first introduced in the Iterated Width algorithm (IW) \cite{Lipovetzky2012IW,Geffner2015GVGIW}. In IW, novelty tests are used for pruning in Breadth-First Search (BrFS). Whenever a state $s$ is generated in a BrFS, a novelty measure (described in more detail below) $nov(s)$ is computed. This is a measure of the extent to which $s$ is ``new'' with respect to all previously generated states. States with a lower measure are ``more novel'' than states with a higher measure \cite{Lipovetzky2012IW}. The original IW algorithm consists of a sequence of calls to IW($0$), IW($1$), etc., where IW($i$) is a BrFS that prunes a state $s$ if $nov(s) > i$. In GVGP, it was found that it is only feasible to run a single IW($i$) iteration \cite{Geffner2015GVGIW}. The best results were obtained with IW($1$), and a variant named IW($\frac{3}{2}$) (see \cite{Geffner2015GVGIW} for details).

The definition of the novelty measure $nov(s)$ of a state $s$ requires $s$ to be defined in terms of a set of boolean features. An example of a boolean feature that can be a part of a state is a predicate \textit{at(cell, type)}, which is true in $s$ if and only if there is an object of the given type in the given cell in $s$. Then, $nov(s)$ is defined as the size of the smallest tuple of features that are all true in $s$, and not all true in any other state generated previously in the same search process. If there is no such tuple, $s$ must be an exact copy of some previously generated state, and $nov(s)$ is defined as $n + 1$, where $n$ is the number of features that are defined. For example, suppose that in $s$, $at((x, y), i) = true$, and in all previously generated states, $at((x, y), i) = false$. Then, $nov(s) = 1$, because there is a tuple of size $1$ of features that were not all true in any previously generated state.

IW($1$) prunes any state $s$ with $nov(s) > 1$. In this paper, \textit{Novelty-Based Pruning} (NBP) is proposed as an idea to prune nodes based on novelty tests in MCTS. The goal is not to prune \textit{bad} lines of play, but to prune \textit{redundant} lines of play.

MCTS often generates states deep in the tree before other states close to the root. For instance, the last state of the first play-out is much deeper in the tree than the first state of the second play-out. This is an important difference with the BrFS used by IW. It means that the novelty measure $nov(s)$ of a state $s$ should be redefined in such a way that it not necessarily uses \textit{all} previously generated states, but only a specific set of states, referred to as the \textit{neighborhood} $N(s)$ of $s$.

\begin{figure}[t]
\centering
\includegraphics[scale=0.4]{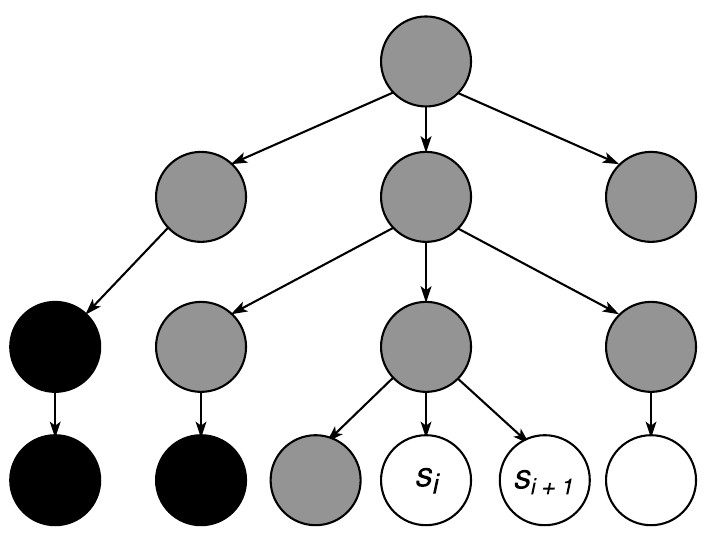}
\vspace{-12pt}
\caption{States used for NBP in MCTS. The grey states are the neighborhood of $s_i$ in MCTS. For $s_{i + 1}$, $s_i$ is also included. The black states would be included for the novelty tests in IW, but not in MCTS.}
\vspace{-15pt}
\label{fig:NBP_MCTS}
\end{figure}

$N(s)$ is the union of four sets of states. The first set consists of the siblings on the ``left'' side of $s$. The ordering of the states matters, but can be arbitrary (as in a BrFS). The second set contains only the parent $p(s)$ of $s$. The third set consists of \textit{all} siblings of $p(s)$. The fourth set is the neighborhood of $p(s)$. More formally, let $s_i$ denote the $i^{th}$ successor of a parent $p(s_i)$. Then, $N(s_i)$ is defined as $N(s_i) = \{s_1, s_2, \dots, s_{i - 1}\} \cup \{p(s_i)\} \cup Sib(p(s_i)) \cup N(p(s_i))$, where $Sib(p(s_i))$ denotes the set of siblings of $p(s_i)$. For the root state $r$, $N(r) = Sib(r) = \varnothing$. An example is depicted in \reffigure{fig:NBP_MCTS}. 

Using the above definition of $N(s)$, $nov(s, N(s))$ is defined as the size of the smallest tuple of features that are all true in $s$, and not all true in any other state in the set $N(s)$. The novelty tests are used in MCTS as follows. Let $n$ be a node with a list of successors $Succ(n)$. The first time that the selection step reaches $n$ when it is fully expanded, all successors $Succ(n)$ are novelty tested based on a single state generated per node, using a threshold of $1$ for the novelty tests (as in IW($1$)). The same boolean features are used to define states in GVG-AI as described in \cite{Geffner2015GVGIW}. Nodes are marked as not being novel if they fail the novelty test. Whenever \textit{all} successors of a node are marked as not novel, that node itself is also marked as not novel. There are a few exceptions where nodes are not marked. If a state has a higher game score than the parent, it is always considered to be novel. Additionally, states transitioned into by playing a movement action are always considered to be novel in games where either only horizontal, or only vertical movement is available (because these games often require moving back and forth which can get incorrectly pruned by NBP otherwise), and in games where the avatar has a movement speed $\leq 0.5$ (because slow movement does not result in the avatar reaching a new cell every tick, and is therefore not detected by the cell-based boolean features).

In the selection step of MCTS, when one of the successors $Succ(n)$ of $n$ should be selected, any successor $n' \in Succ(n)$ is ignored if it is marked as not novel, unless the average normalized score $\mean{Q}(n) < 0.5$. In such cases, the situation is considered to be dangerous and all alternatives should be considered to see if a better position can be found. For the final selection of the move to play in the real game, non-novel nodes are also only considered if the best novel alternative has a normalized average score $< 0.5$.

When the successors $Succ(n)$ have been novelty tested, every node $n_i \in Succ(n)$ stores a set of tuples of features that were all true in the states generated for the purpose of novelty testing for the nodes $\{n\} \cup Succ(n)$. This means that the tuples of features that are true in the neighborhood $N(s)$ of a state $s$ can be reconstructed relatively efficiently by traversing the path from $s$ back to the root, and collecting the tuples in the stored sets. This is the main reason for defining $N(s)$ as described above. Including more states (for instance, the black states in \reffigure{fig:NBP_MCTS}) would require also traversing back down the tree to collect more sets of tuples. This could increase the number of nodes that NBP marks as not being novel, but would also be more expensive computationally. This is not a problem in the BrFS of IW, because it can simply store all tuples of features that are all true in any generated state in the same set for the entire search process.

Novelty measures are assigned to nodes based on only one state per node. Therefore, given two identical open-loop game trees in nondeterministic games, it is possible that a node in one tree is pruned and the equivalent node in the other tree is not pruned. For this reason, when combining NBP with Tree Reuse, the results of novelty tests on nodes in the first ply below the new root node are reset when reusing the previous tree. This does not entirely remove the influence of nondeterminism on NBP, but close to the root that influence is at least reduced.

\subsection{Knowledge-Based Evaluations}
An important problem with MCTS in GVG-AI is that it is often infeasible to find any terminal states, or even states with a change in game score. This means that the evaluation function in \refequation{Eq:GVGEval} often returns the same value for all states generated in the same tick, and MCTS explores the search space and behaves randomly. In this paper, a heuristic evaluation function is proposed that uses knowledge collected during simulations, and distances to objects that could potentially be interesting, to distinguish between states that have identical evaluations according to \refequation{Eq:GVGEval}. The basic idea is not new; some agents in the competition of 2014 used distance-based evaluation functions \cite{Perez2015GVG14}. A similar idea is also described in \cite{Perez2014KB}, and extended in \cite{Eeden2015Thesis}. The idea discussed here is based on the same intuition, but a number of implementation details are different. Another related idea is described in \cite{Chu2015PathfindingMCTSGVGP}, where MCTS is used to learn which objects are interesting, and a pathfinding algorithm is used to move towards a selected goal.

Let $X(s_0)$ denote the evaluation of the current game state $s_0$, and let $X(s_T)$ denote the evaluation of the final state $s_T$ of a play-out. If $X(s_T) = X(s_0)$, a heuristic evaluation $Eval_{KB}(s_T)$ is computed and added to $X(s_T)$. For every object type $i$ observed in a game, let $d_0(i)$ denote the distance from the avatar to the closest object of type $i$ in $s_0$, and let $d_T(i)$ denote the distance from the avatar to the closest object of type $i$ in $s_T$. These distances are computed using the A* pathfinding algorithm \cite{Hart1968AStar}. The pathfinding algorithm takes objects of the \textit{wall} type into account as obstacles. Many games can also contain other objects that block movement, or portals that can be used for teleportation. These objects are not taken into account, because the agent would first need to learn how these objects influence pathfinding. For every object type $i$, a weight $w_i$ is used to reward or punish the agent for moving to objects of that type. This is done by computing $Eval_{KB}(s_T)$ as given by \refequation{Eq:EvalKB}, normalizing it to lie in $[0, 0.5]$, and adding it to $X(s_T)$ if otherwise $X(s_T) = X(s_0)$.
\begin{equation}
Eval_{KB}(s_T) = \sum_i w_i \times (d_0(i) - d_T(i))
\label{Eq:EvalKB}
\end{equation}
Object types $i$ with a small absolute weight ($|w_i| < 10^{-4}$) are ignored, to save the computational cost of pathfinding.

The weights $w_i$ are determined as follows. To motivate exploration, all weights are initialized with positive values ($0.1$ for NPCs, $0.25$ for Movables, and $1$ for Resources and Portals), and incremented by $10^{-4}$ every game tick. States $s_t$ generated during the selection or play-out steps of MCTS are used to adjust these weights. Let $s_{t - 1}$ denote the predecessor of $s_t$. Whenever such a state $s_t$ is generated, it is used to update some of the weights $w_i$. The intuition is that, if $X(s_t) \neq X(s_{t - 1})$, it is likely that some interesting collision event occurred in the transition from $s_{t - 1}$ to $s_t$ that caused the change in score. The framework provides access to a set $E(s_t)$ of collision events that occurred in that transition. Every event $e \in E(s_t)$ is a collision event between two objects, where one object is either the avatar, or an object created by the avatar (for instance, a missile fired by the avatar), and the other object is of some type $i$. Let $\Delta = X(s_t) - X(s_{t - 1})$ denote the observed change in score. For every object type $i$, a sum $\Delta_i$ is kept of all changes in scores observed in state transitions where collision events with objects of type $i$ occurred. Additionally, a counter $n_i$ of event occurrences is kept for every type $i$, such that the average change in score $\mean{\Delta}_i = \frac{\Delta_i}{n_i}$ for collisions with every type can be computed. Whenever an event with an object of type $i$ is observed, $w_i$ is updated as given by Formula \ref{Eq:KBWeightUpdate}.
\begin{equation}
w_i \gets w_i + (\mean{\Delta}_i - w_i) \times \alpha_i
\label{Eq:KBWeightUpdate}
\end{equation}
$\alpha_i$ is a learning rate that is initialized to $0.8$ for every type, and updated as given by Formula \ref{Eq:KBAlphaUpdate} after updating $w_i$.
\begin{equation}
\alpha_i \gets \max(0.1, 0.75 \times \alpha_i)
\label{Eq:KBAlphaUpdate}
\end{equation}
This idea is similar to using gradient descent for minimizing $|\mean{\Delta}_i - w_i|$. The main reason for not simply using $\mean{\Delta}_i$ directly is to avoid relying too much on the knowledge obtained from a low number of observed events.

\subsection{Deterministic Game Detection}
The idea of \textit{Deterministic Game Detection} (DGD) is to detect when a game is likely to be deterministic, and treat deterministic games differently from nondeterministic games. At the start of every game, $M$ random sequences of actions of length $N$ are generated. Each of the $M$ sequences is used to advance a copy of the initial game state $s_0$, with $R$ repetitions per sequence. If any of the $M$ action sequences did not result in equivalent states among the $R$ repetitions for that sequence, the game is classified as nondeterministic. Additionally, any game in which NPCs are observed is immediately classified as nondeterministic. Any other game is classified as deterministic. In this paper, $M = N = 5$ and $R = 3$.

Many participants in previous GVG-AI competitions \cite{PerezGVG15} used a similar idea to switch to a different algorithm for deterministic games (for instance, using Breadth-First Search in deterministic games and MCTS in nondeterministic games). In this paper, DGD is only used to modify MCTS and the TR and NBP enhancements in deterministic games. The $\mean{Q}(S_i)$ term in \refequation{Eq:UCT} (or the equivalent term in the formula of PH) is replaced by $\frac{3}{4} \times \mean{Q}(S_i) + \frac{1}{4} \times \hat{Q}_{max}(S_i)$, where $\hat{Q}_{max}(S_i)$ is the maximum score observed in the subtree rooted in $S_i$. This is referred to as \textit{mixmax} \cite{Jacobsen2014MCTSMario,Frydenberg2015MCTSGVGP}. Additionally, TR and NBP are modified to no longer decay or reset any old results.

\section{Experiments} \label{sec:Experiments}
\subsection{Setup}
The enhancements discussed in this paper have been experimentally evaluated using the following setup. Every experiment was run using six sets that are available in the framework, of ten games each, for a total of sixty different games per experiment. \reftable{TableGameSets} lists the names of the games for every set. Average results are presented for every set of games, and for the total of all sixty games combined. For every game, five different levels were used, with a minimum of fifteen repetitions per level per experiment (leading to a minimum of 750 runs per set). 95\% confidence intervals are presented for all results. All games were played according to the GVG-AI competition rules\footnote{Revision \scriptsize{24b11aea75722ab02954c326357949b97efb7789} \footnotesize{of the GVG-AI framework (https://github.com/EssexUniversityMCTS/gvgai) was used.}}, on a CentOS Linux server consisting of four AMD Twelve-Core OpteronT 6174 processors (2.2 GHz).

\subsection{Results}
\begin{table}[t]
\footnotesize
\renewcommand{\arraystretch}{1.2}
\caption{Win Percentages (Benchmark Agents, 1000 runs per set)}
\vspace{-5pt}
\label{TableBaselines}
\centering
\begin{tabular}{|c|cccc|}
\hline
Sets & SOLMCTS & MCTS & IW(1) & \textsc{YBCriber} \\
\hline
Set 1 & 34.5 $\pm$ 2.9 & 41.7 $\pm$ 3.1 & 55.8 $\pm$ 3.1 & 68.8 $\pm$ 2.9 \\
Set 2 & 33.4 $\pm$ 2.9 & 33.5 $\pm$ 2.9 & 47.0 $\pm$ 3.1 & 65.0 $\pm$ 3.0 \\
Set 3 & 13.2 $\pm$ 2.1 & 23.0 $\pm$ 2.6 & 17.8 $\pm$ 2.4 & 40.3 $\pm$ 3.0 \\
Set 4 & 28.3 $\pm$ 2.8 & 30.5 $\pm$ 2.9 & 30.6 $\pm$ 2.9 & 43.5 $\pm$ 3.1 \\
Set 5 & 19.7 $\pm$ 2.5 & 28.9 $\pm$ 2.8 & 17.5 $\pm$ 2.4 & 42.6 $\pm$ 3.1 \\
Set 6 & 30.1 $\pm$ 2.8 & 28.6 $\pm$ 2.8 & 32.8 $\pm$ 2.9 & 54.4 $\pm$ 3.1 \\
\hline
\hline
Total & 26.5 $\pm$ 1.1 & 31.0 $\pm$ 1.2 & 33.6 $\pm$ 1.2 & 52.4 $\pm$ 1.3 \\
\hline
\end{tabular}
\vspace{-17pt}
\end{table}
In the first experiment, the following benchmark agents are compared to each other; SOLMCTS, MCTS, IW(1), and \textsc{YBCriber}. SOLMCTS is the Sample Open Loop MCTS controller included in the framework. MCTS is our baseline implementation of MCTS, based on the \textsc{MaastCTS} \cite{Schuster2015Thesis} agent, which has a number of differences in comparison to SOLMCTS. MCTS expands all nodes for states generated in simulations (as opposed to one node per simulation), $C$ is set to $0.6$ in the UCB1 equation (as opposed to $C = \sqrt{2}$), it simulates up to ten actions after the selection step (as opposed to ten steps from the root node), it uses the 1 second of initialization time for running the algorithm (as opposed to not using that time), and it plays the action with the maximum average score (as opposed to the maximum visit count). IW($1$) is the Iterated Width-based agent, as described in \cite{Geffner2015GVGIW}. \textsc{YBCriber} is an IW-based agent with a number of other features, which won the GVG-AI competition at the IEEE CEEC 2015 conference. The results are given in \reftable{TableBaselines}. The experimental data reveals that the baseline MCTS agent outperforms SOLMCTS. IW(1) performs slightly better than MCTS overall, and \textsc{YBCriber} performs much better than the other benchmark agents.

In \reftable{TableBFTI}, our MCTS implementation with Breadth-First Tree Initialization and Safety Prepruning (BFTI) is compared to the MCTS implementation without BFTI. The results for MCTS are based on 1000 runs per set, and the results for BFTI on 750 runs per set. BFTI appears to lower the win percentage slightly, but the 95\% confidence intervals overlap. The two columns on the right-hand side show the percentage of lost games where the game was terminated before $t = 2000$ (where $t = 2000$ is the maximum duration of a game in GVG-AI). BFTI reduces this percentage significantly. Even though it may slightly decrease win percentages, the quality of play in lost games can be considered to be improved; the agent delays a significant number of losses. This may leave more time for other enhancements to find wins. Therefore, BFTI is included in the baseline MCTS agent for the following experiments that evaluate other enhancements individually. This is followed by an experiment with more enhancements combined.
\begin{table}[t]
\footnotesize
\renewcommand{\arraystretch}{1.1}
\caption{Breadth-First Tree Initialization with Safety Prepruning}
\vspace{-8pt}
\label{TableBFTI}
\centering
\begin{tabular}{|c|cc|cc|}
\hline
& \multicolumn{2}{c}{Win Percentage} \vrule & \multicolumn{2}{c}{\% of Losses $t<2000$} \vrule \\
Sets & BFTI & MCTS & BFTI & MCTS \\
\hline
Set 1 & 43.3 $\pm$ 3.5 & 41.7 $\pm$ 3.1 & 42.6 $\pm$ 4.7 & 52.8 $\pm$ 4.1 \\
Set 2 & 33.1 $\pm$ 3.4 & 33.5 $\pm$ 2.9 & 50.8 $\pm$ 4.4 & 51.1 $\pm$ 3.8 \\
Set 3 & 21.2 $\pm$ 2.9 & 23.0 $\pm$ 2.6 & 0.0 $\pm$ 0.0 & 16.1 $\pm$ 2.6 \\
Set 4 & 30.3 $\pm$ 3.3 & 30.5 $\pm$ 2.9 & 73.4 $\pm$ 3.8 & 76.8 $\pm$ 3.1 \\
Set 5 & 23.1 $\pm$ 3.0 & 28.9 $\pm$ 2.8 & 72.4 $\pm$ 3.6 & 73.7 $\pm$ 3.2 \\
Set 6 & 29.2 $\pm$ 3.3 & 28.6 $\pm$ 2.8 & 69.3 $\pm$ 3.9 & 72.3 $\pm$ 3.3 \\
\hline
\hline
Total & 30.0 $\pm$ 1.3 & 31.0 $\pm$ 1.2 & 51.0 $\pm$ 1.7 & 56.7 $\pm$ 1.5 \\
\hline
\end{tabular}
\vspace{-12pt}
\end{table}

\reftable{TableProgHistNST} shows the win percentages obtained by adding Progressive History (PH), N-Gram Selection Technique (NST), or both to the BFTI agent. PH and NST appear to increase the average win percentage, but the confidence intervals overlap. The two combined result in a statistically significant increase.
\begin{table}[t]
\footnotesize
\renewcommand{\arraystretch}{1.2}
\caption{Win Percentages (PH and NST, 750 runs per set)}
\vspace{-10pt}
\label{TableProgHistNST}
\centering
\begin{tabular}{|c|cccc|}
\hline
Sets & BFTI & PH & NST & NST+PH \\
\hline
Set 1 & 43.3 $\pm$ 3.5 & 43.2 $\pm$ 3.5 & 45.1 $\pm$ 3.6 & 43.5 $\pm$ 3.5 \\
Set 2 & 33.1 $\pm$ 3.4 & 34.5 $\pm$ 3.4 & 36.5 $\pm$ 3.4 & 38.0 $\pm$ 3.5 \\
Set 3 & 21.2 $\pm$ 2.9 & 23.3 $\pm$ 3.0 & 23.1 $\pm$ 3.0 & 24.1 $\pm$ 3.1 \\
Set 4 & 30.3 $\pm$ 3.3 & 29.5 $\pm$ 3.3 & 29.7 $\pm$ 3.3 & 32.3 $\pm$ 3.3 \\
Set 5 & 23.1 $\pm$ 3.0 & 23.9 $\pm$ 3.1 & 30.0 $\pm$ 3.3 & 28.0 $\pm$ 3.2 \\
Set 6 & 29.2 $\pm$ 3.3 & 30.0 $\pm$ 3.3 & 31.1 $\pm$ 3.3 & 33.1 $\pm$ 3.4 \\
\hline
\hline
Total & 30.0 $\pm$ 1.3 & 30.7 $\pm$ 1.3 & 32.6 $\pm$ 1.4 & 33.2 $\pm$ 1.4 \\
\hline
\end{tabular}
\vspace{-18pt}
\end{table}

\begin{figure}[t]
\centering
\vspace{-5pt}
\includegraphics[scale=0.30]{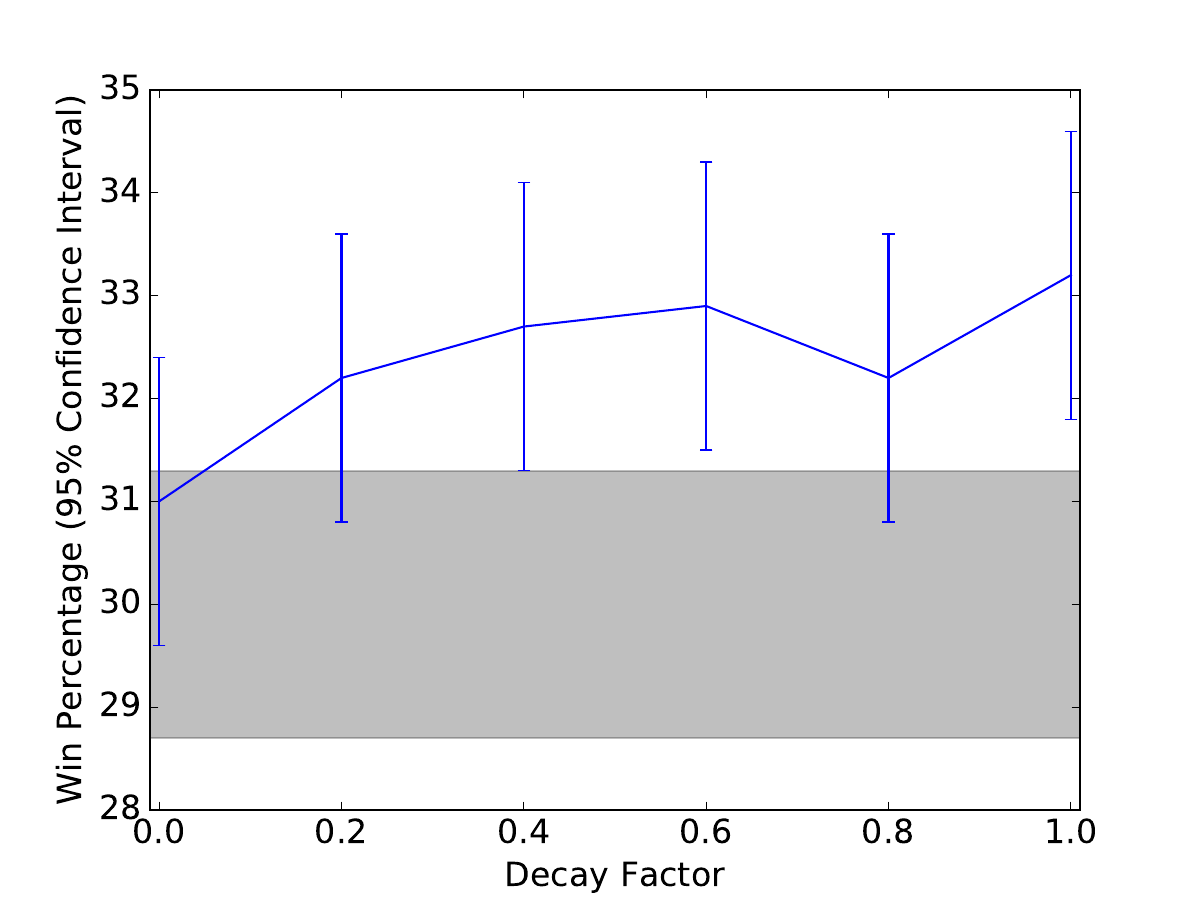}
\vspace{-10pt}
\caption{95\% confidence intervals for win percentages of BFTI with Tree Reuse (TR) for different values of the decay factor $\gamma$. The area shaded in grey is the confidence interval for the win percentage of BFTI without TR.}
\label{fig:WinPercentagesTreeReuse}
\vspace{-12pt}
\end{figure}

\reffigure{fig:WinPercentagesTreeReuse} depicts 95\% confidence intervals for the win percentage of the BFTI agent with Tree Reuse (TR), for six different values of the decay factor $\gamma$. The confidence interval for BFTI is shaded in grey. TR with $\gamma \in \{0.4, 0.6, 1.0\}$ significantly improves the win percentage of BFTI.

\reftable{TableKBELANBP} shows the win percentages of adding either Knowledge-Based Evaluations (KBE), Loss Avoidance (LA) or Novelty-Based Pruning (NBP) to the BFTI agent. All three individually show an increase in the average win percentage over BFTI, with KBE giving the largest increase.
\begin{table}[t]
\footnotesize
\renewcommand{\arraystretch}{1.2}
\caption{Win Percentages (KBE, LA and NBP, 750 runs per set)}
\vspace{-10pt}
\label{TableKBELANBP}
\centering
\begin{tabular}{|c|cccc|}
\hline
Sets & BFTI & KBE & LA & NBP \\
\hline
Set 1 & 43.3 $\pm$ 3.5 & 50.4 $\pm$ 3.6 & 52.0 $\pm$ 3.6 & 49.6 $\pm$ 3.6 \\
Set 2 & 33.1 $\pm$ 3.4 & 52.3 $\pm$ 3.6 & 34.0 $\pm$ 3.4 & 34.5 $\pm$ 3.4 \\
Set 3 & 21.2 $\pm$ 2.9 & 19.1 $\pm$ 2.8 & 23.3 $\pm$ 3.0 & 23.5 $\pm$ 3.0 \\
Set 4 & 30.3 $\pm$ 3.3 & 30.1 $\pm$ 3.3 & 29.6 $\pm$ 3.3 & 32.0 $\pm$ 3.3 \\
Set 5 & 23.1 $\pm$ 3.0 & 31.3 $\pm$ 3.3 & 31.9 $\pm$ 3.3 & 23.9 $\pm$ 3.1 \\
Set 6 & 29.2 $\pm$ 3.3 & 33.2 $\pm$ 3.4 & 28.8 $\pm$ 3.2 & 34.8 $\pm$ 3.4 \\
\hline
\hline
Total & 30.0 $\pm$ 1.3 & 36.1 $\pm$ 1.4 & 33.3 $\pm$ 1.4 & 33.0 $\pm$ 1.4 \\
\hline
\end{tabular}
\vspace{-20pt}
\end{table}

\reftable{TableEnhancementsCombined} shows the win percentages of a number of variants of MCTS with multiple enhancements combined. ``No DGD'' is an agent with all enhancements discussed in this paper, except for Deterministic Game Detection (DGD). ``No BFTI'' is an agent with all enhancements except for BFTI. This is added to test the assumption made earlier that the ability of BFTI to delay games may enable other enhancements to find more wins. The last agent contains all enhancements. In combination with all the other enhancements, DGD significantly improves the win percentage. DGD was found not to provide a significant increase in win percentage when applied to the BFTI, TR ($\gamma = 0.6$) or NBP agents without other enhancements (those results have been omitted to save space). Additionally, BFTI appears to increase the win percentage in combination with all other enhancements, whereas \reftable{TableBFTI} shows it appears to \textit{decrease} the win percentage when other enhancements are absent, but these differences are not statistically significant.
\begin{table}[t]
\footnotesize
\renewcommand{\arraystretch}{1.2}
\caption{Win Percentages (Enhancements Combined, 750 runs per set)}
\vspace{-8pt}
\label{TableEnhancementsCombined}
\centering
\begin{tabular}{|c|cccc|}
\hline
Sets & BFTI & No DGD & No BFTI & All Enhanc. \\
\hline
Set 1 & 43.3 $\pm$ 3.5 & 62.7 $\pm$ 3.5 & 62.7 $\pm$ 3.5 & 62.8 $\pm$ 3.5 \\
Set 2 & 33.1 $\pm$ 3.4 & 56.4 $\pm$ 3.5 & 55.7 $\pm$ 3.6 & 59.3 $\pm$ 3.6 \\
Set 3 & 21.2 $\pm$ 2.9 & 22.1 $\pm$ 3.0 & 28.5 $\pm$ 3.2 & 28.7 $\pm$ 3.2 \\
Set 4 & 30.3 $\pm$ 3.3 & 32.7 $\pm$ 3.4 & 47.1 $\pm$ 3.6 & 48.1 $\pm$ 3.6 \\
Set 5 & 23.1 $\pm$ 3.0 & 37.2 $\pm$ 3.5 & 39.6 $\pm$ 3.5 & 42.1 $\pm$ 3.5 \\
Set 6 & 29.2 $\pm$ 3.3 & 38.3 $\pm$ 3.5 & 49.2 $\pm$ 3.6 & 49.1 $\pm$ 3.6 \\
\hline
\hline
Total & 30.0 $\pm$ 1.3 & 41.6 $\pm$ 1.4 & 47.1 $\pm$ 1.5 & 48.4 $\pm$ 1.5 \\
\hline
\end{tabular}
\vspace{-12pt}
\end{table}

\section{Conclusion and Future Work} \label{sec:Conclusion}
Eight enhancements for Monte-Carlo Tree Search (MCTS) in General Video Game Playing (GVGP) have been discussed and evaluated. Most of them have been shown to significantly (95\% confidence) increase the average win percentage over sixty different games when added individually to MCTS. All the enhancements combined increase the win percentage of our basic MCTS implementation from $31.0 \pm 1.2$ to $48.4 \pm 1.5$. This final performance is relatively close to the win percentage of the winner of the IEEE CEEC 2015 conference; \textsc{YBCriber}, with a win percentage of $52.4 \pm 1.3$.

Many of the discussed enhancements have parameters, which so far have only been tuned according to short, preliminary experiments. These parameters can likely be tuned better in future work to improve the performance. Loss Avoidance (LA) and Novelty-Based Pruning (NBP) as proposed in this paper have binary effects, in that LA backpropagates only one result from multiple generated siblings and NBP classifies nodes as either novel or not novel. Perhaps these can be improved by making them less binary. The overall performance of the agent can also likely be improved by incorporating more features that are commonly seen among the top entries in past competitions, such as the use of influence maps \cite{Millington2009AIForGames}. Finally, some of the new enhancements for MCTS, such as LA and NBP, can be evaluated in domains other than GVG-AI.

\begin{table}[t]
\footnotesize
\renewcommand{\arraystretch}{1.0}
\caption{Names of the Games in every Set}
\vspace{-8pt}
\label{TableGameSets}
\centering
\begin{tabular}{|c|c|}
\hline
Set 1 & \makecell{\textit{\scriptsize{Aliens, Boulderdash, Butterflies, Chase, Frogs, Missile}}\\ \textit{\scriptsize{Command, Portals, Sokoban, Survive Zombies, Zelda}}} \\
\hline
Set 2 & \makecell{\textit{\scriptsize{Camel Race, Digdug, Firestorms, Infection, Firecaster,}}\\ \textit{\scriptsize{Overload, Pacman, Seaquest, Whackamole, Eggomania}}} \\
\hline
Set 3 & \makecell{\textit{\scriptsize{Bait, BoloAdventures, BrainMan, ChipsChallenge, Modality,}}\\ \textit{\scriptsize{Painter, RealPortals, RealSokoban, TheCitadel, ZenPuzzle}}} \\
\hline
Set 4 & \makecell{\textit{\scriptsize{Roguelike, Surround, Catapults, Plants, Plaque-Attack,}}\\ \textit{\scriptsize{Jaws, Labyrinth, Boulderchase, Escape, Lemmings}}} \\
\hline
Set 5 & \makecell{\textit{\scriptsize{Solarfox, Defender, Enemy Citadel, Crossfire, Lasers,}}\\ \textit{\scriptsize{Sheriff, Chopper, Superman, WaitForBreakfast, CakyBaky}}} \\
\hline
Set 6 & \makecell{\textit{\scriptsize{Lasers 2, Hungry Birds, Cook me Pasta, Factory Manager, Race}}\\ \textit{\scriptsize{Bet 2, Intersection, Black Smoke, Ice and Fire, Gymkhana, Tercio}}} \\
\hline
\end{tabular}
\vspace{-13pt}
\end{table}

\section*{Acknowledgement}
This work is partially funded by the Netherlands Organisation for Scientific Research (NWO) in the framework of the project GoGeneral, grant number 612.001.121.

\bibliographystyle{IEEEtran}
\bibliography{IEEEabrv,ReferencesCIG}

\end{document}